\documentclass[10pt,twocolumn,letterpaper]{article}

\usepackage[pagenumbers]{wacv} 

\usepackage{booktabs}
\usepackage{graphicx}
\usepackage{subcaption}
\usepackage{comment}
\usepackage{url}
\usepackage{siunitx}
\usepackage[accsupp]{axessibility}  

\newcommand{\modelfull}{Spatio-Temporal Gaze Network }
\newcommand{\model}{ST-Gaze}

\definecolor{wacvblue}{rgb}{0.21,0.49,0.74}
\usepackage[pagebackref,breaklinks,colorlinks,allcolors=wacvblue]{hyperref}

\def\wacvPaperID{726} 
\def\confName{WACV}
\def\confYear{2026}

\title{Learning spatio-temporal feature representations for video-based gaze estimation}

\author{Alexandre Personnic, Mihai Bâce
\vspace{0.1cm}
\\
KU Leuven, Department of Computer Science, Group T Leuven Campus, Leuven, Belgium\\
{\tt\small \{alexandre.personnic, mihai.bace\}@kuleuven.be}
}

\usepackage{xcolor} 

\begin{document}
\maketitle
\begin{abstract}
    Video-based gaze estimation methods aim to capture the inherently temporal dynamics of human eye gaze from multiple image frames. 
    However, since models must capture both spatial and temporal relationships, performance is limited by the feature representations within a frame but also between multiple frames. 
    We propose the \modelfull{} (\model{}), a model that combines a CNN backbone with dedicated channel attention and self-attention modules to fuse eye and face features optimally. 
    The fused features are then treated as a spatial sequence, allowing for the capture of an intra-frame context, which is then propagated through time to model inter-frame dynamics. 
    We evaluated our method on the EVE dataset and show that \model{} achieves state-of-the-art performance both with and without person-specific adaptation. 
    Additionally, our ablation study provides further insights into the model performance, showing that preserving and modelling intra-frame spatial context with our spatio-temporal recurrence is fundamentally superior to premature spatial pooling. 
    As such, our results pave the way towards more robust video-based gaze estimation using commonly available cameras. 
\end{abstract}    
\section{Introduction}
\label{sec:intro}

Gaze behaviour is a powerful indicator of human intention and attention \cite{shepherdFollowingGazeGazeFollowing2010}, making gaze estimation a valuable tool in domains ranging from interactive systems like HCI and gaming \cite{majarantaEyeTrackingEyeBased2014, petersModelAttentionInterest2005, heIntegratingEyeTrackingGames2024} to analytical applications in healthcare and education \cite{wasfyEyeingInterfaceAdvancing2024, tonoApplicationEyeTrackingEfl2011, yamamotoTeachersGazeAwareness2013, harezlakApplicationEyeTracking2018}.
To make this technology widely accessible, recent research has focused on appearance-based methods that leverage deep learning on commodity cameras, supplanting traditional model-based techniques \cite{hansenEyeBeholderSurvey2010}. The success of mapping static images to gaze directions \cite{zhangAppearancebasedGazeEstimation2015a, akinyeluConvolutionalNeuralNetworkBased2020, pathiranaEyeGazeEstimation2022} has naturally led to video-based estimation, where Recurrent Neural Networks (RNNs) are used to model the temporal dynamics of eye movements, further improving robustness and accuracy \cite{parkEndtoendVideobasedEyeTracking2020, renGazeEstimationBased2023, chengAppearancebasedGazeEstimation2024}.

Despite significant progress in recent years, the performance of appearance-based models is often limited by two fundamental challenges. 
The first is the presence of unobservable, person-specific anatomical factors, such as the angle difference between the optical and visual axis of the eye, commonly referred to as the kappa angle \cite{guestrinGeneralTheoryRemote2006}.
This creates a subject-dependent offset that is difficult to model from images alone \cite{parkFewShotAdaptiveGaze2019a, jinKappaAngleRegression2023}.
To tackle this challenge, several person-specific adaptation and calibration techniques have been studied.
Classical approaches often rely on an explicit calibration phase, where a user looks at one or more known points on a screen to map their gaze \cite{guestrinGeneralTheoryRemote2006}.
More recently, learning-based methods have emerged that can adapt a general model to a new person using only a few labelled or unlabelled samples \cite{baoStoryYourEyes2021, parkFewShotAdaptiveGaze2019a, lindenLearningPersonalizeAppearanceBased2019}.
In our work, we evaluate our model's feature representation by using it as a backbone for the Self-Calibration and Person-specific Transform (SCPT) modules proposed by Bao et al. \cite{baoStoryYourEyes2021}.
We chose SCPT as it represents the state-of-the-art in few-shot adaptation on the EVE dataset and does not require labelled calibration data, aligning with our goal of unconstrained gaze estimation.
Our experiments demonstrate that our proposed video-based gaze estimation model achieves state-of-the-art performance when these downstream adaptation techniques are used.

The second, and the focus of our work due to its broad implications for general-purpose video-based gaze estimation, is the quality of feature representations extracted from the input images.
Recent work has demonstrated that incorporating temporal context from video sequences can improve accuracy \cite{parkEndtoendVideobasedEyeTracking2020, renGazeEstimationBased2023}.
However, we hypothesize that the performance of the temporal models is fundamentally constrained by the expressiveness of the static, per-frame features they receive as input. 
Our empirical analyses show that architectural choices in the feature extraction stage, such as the backbone and the fusion mechanism, have a significant impact on overall model performance and are necessary to fully capture the spatial and temporal relationships.

To address this challenge, we propose \textbf{\model{}}, a novel architecture designed to generate a more context-aware feature representation for temporal gaze estimation. \model{} is built on three core principles. 
First, it leverages a convolutional backbone (EfficientNetB3 \cite{tanEfficientNetRethinkingModel2020}) to extract rich initial features from separate eye and face patches. 
The eyes provide the primary signal for gaze direction, while the face offers complementary context, such as head-pose and orientation, which helps disambiguate gaze \cite{baoAdaptiveFeatureFusion2021}.
Second, we propose a combination of attention mechanisms: a channel-wise attention ECA module \cite{wangECANetEfficientChannel2020} adaptively weighs the importance of eye versus face features, while a following self-attention module \cite{vaswaniAttentionAllYou2017} learns the spatial relationships within the fused feature map. 
Finally, we introduce a novel recurrence mechanism to address a critical limitation in prior temporal models: \textit{premature spatial pooling}. 
By collapsing the rich 2D feature map into a single vector before temporal processing, these models irrevocably discard the intra-frame spatial relationships between facial features. 
Our key innovation is a novel \textit{spatio-temporal recurrence} that models this crucial \textit{intra-frame} context first. 
We treat the 2D feature map as a spatial sequence and process it with a Gated Recurrent Unit (GRU) \cite{choLearningPhraseRepresentations2014} to generate a context-aware summary of the frame. 
This summary, in the form of the GRU's final hidden state, is then propagated through time to model the \textit{inter-frame} dynamics. 
This design ensures that temporal modelling operates on a spatially informed representation, a capability our ablation study confirms is critical for performance.

We evaluated our proposed \model{} on the EVE dataset \cite{parkEndtoendVideobasedEyeTracking2020}.
As our work targets video-based gaze estimation for common interactive settings such as desktop and laptop use, EVE is the premier benchmark for this task due to its high-quality, continuous video data of natural user-screen interaction. 
While other large datasets exist, such as the image-based ETH-XGaze \cite{zhangETHXGazeLargeScale2020a} or the fully "in-the-wild" GAZE360 \cite{kellnhoferGaze360PhysicallyUnconstrained2019}, EVE's specific focus aligns directly with our research goals.
\model{} establishes a new state-of-the-art for single-view, video-based 3D gaze estimation on EVE with an error of $2.58^\circ$. 
We limited ourselves to camera-only inputs as, while using screen content can further disambiguate gaze targets \cite{parkEndtoendVideobasedEyeTracking2020}, we aim to create a more general-purpose 3D gaze estimation model that does not require continuous screen recording.
Furthermore, when our model is used as the input to the person-specific adaptation modules of Bao et al. \cite{baoStoryYourEyes2021}, we reduce the final error to $1.87^\circ$ in the offline setting, where all of a person's data is available for calibration, demonstrating the broad utility of our approach.

In summary, our key contributions are:
\begin{itemize}
    \item We propose \model{}, a novel gaze estimation architecture featuring a spatio-temporal recurrence mechanism that models intra-frame spatial context before propagating information temporally to capture inter-frame context.
    \item We achieve state-of-the-art performance on the EVE benchmark for video-based 3D gaze estimation, and show that our model serves as a superior backbone for downstream person-specific adaptation methods.
    \item We provide a comprehensive ablation study that disentangles the contributions of our design choices, including the backbone, attention mechanisms, and the spatio-temporal recurrence, revealing that our proposed recurrence structure and the self-attention module are the most critical drivers of performance.
\end{itemize}
A public version of our repository can be found in our institutional GitLab at the following address:
\url{https://gitlab.kuleuven.be/u0172623/ST-Gaze}.
\section{Related Work}
\label{sec:rel}

\paragraph{Appearance-based Gaze Estimation.} 
This approach aims to directly learn a mapping from images of the face or eyes to a gaze direction, bypassing explicit 3D modelling. 
Early work demonstrated the feasibility of this approach \cite{zhangAppearancebasedGazeEstimation2015a}, and the introduction of large-scale datasets like MPIIGaze \cite{zhangMPIIGazeRealWorldDataset2019}, ETH-XGaze \cite{zhangETHXGazeLargeScale2020a} or EVE \cite{parkEndtoendVideobasedEyeTracking2020} spurred the development of deep learning solutions. 
Many methods have explored the optimal source of image information, with some using only eye patches and others arguing for the inclusion of the full face to provide head pose context \cite{zhangItsWrittenAll2017}. 
A common theme is the fusion of features from multiple streams, such as left and right eye patches, or eyes and the face. 
However, this fusion is often a simple concatenation followed by a fully-connected layer. 
More recent works \cite{baoAdaptiveFeatureFusion2021, renGazeEstimationBased2023, liAppearanceBasedGazeEstimation2023a} proposed different ways to more intelligently combine eye features, demonstrating that a more sophisticated fusion strategy can yield significant gains. 
Our work extends this idea by using both channel and self-attention for a more comprehensive fusion of eye and face features

\paragraph{Temporal Models in Gaze Estimation.}
Gaze is inherently a dynamic process. 
Recognising this, several works have moved from single-image estimation to using video sequences. 
The work of Park et al. \cite{parkEndtoendVideobasedEyeTracking2020} introduced the EVE dataset and demonstrated that a GRU-based recurrent model (EyeNet-GRU) could leverage temporal consistency to improve upon a static baseline. 
Similarly, Zhou et al. \cite{zhouLearning3DGaze2019} combined an iTracker-style model with a bidirectional LSTM to capture temporal dynamics. 
Li et al. \cite{liAppearanceBasedGazeEstimation2023a} proposed a Static Transformer Temporal Differential Network (STTDN) which uses a custom RNN cell to extract sight movement. 
Ren et al. \cite{renGazeEstimationBased2023} also incorporated a GRU to learn eye movement dynamics.
These works firmly establish the value of temporal information.
However, they are all dependent on the quality of the feature vector extracted at each time step. Moreover, they are spatially pooling the features before the temporal module.
In contrast to these works which pool spatial features into a single vector, our approach preserves the full spatial feature map. 
We argue this is critical, as it allows our model to reason about both intra-frame spatial relationships and inter-frame temporal dynamics, a capability lost in previous methods.

\paragraph{Attention and Transformers for Gaze Estimation.}
With their success in other vision tasks \cite{pereiraReviewTransformerBasedModels2024}, attention mechanisms and Transformer architectures are emerging in gaze estimation.
Cheng and Lu \cite{chengGazeEstimationUsing2021} were among the first to explore a hybrid Vision Transformer (ViT) model.
Crucially, their work demonstrated that a hybrid approach, which uses a CNN to extract local feature maps before a Transformer models global relationships, outperforms both pure CNN and pure ViT architectures.
Nagpure and Okuma \cite{nagpureSearchingEfficientNeural2023} used neural architecture search to find an efficient Transformer-based model.
More closely related to our work, Ren et al. \cite{renGazeEstimationBased2023} integrated channel and self-attention modules to enhance feature extraction from fused eye and face features.
Our \model{} builds on this but makes a critical architectural distinction.
While existing methods typically apply attention and then pool the features to create a single vector for the temporal model, we innovate by preserving the spatial sequence from our self-attention module. 
By applying a GRU first across this spatial sequence (intra-frame) and then propagating the hidden state across time (inter-frame), we create a recurrent structure that better models the complex spatio-temporal nature of gaze.
This approach draws inspiration from architectures in other video understanding tasks, such as ConvGRU \cite{ballasDelvingDeeperConvolutional2016}, and broader tasks tackling spatio-temporal sequences \cite{liuSpatioTemporalGRUTrajectory2019}.
Our specific approach is, to the best of our knowledge, novel in the context of gaze estimation.
Finally, our work also connects to the line of research on person-specific modelling \cite{baoStoryYourEyes2021, parkFewShotAdaptiveGaze2019a}.
By providing a much stronger baseline model (\model{}), we show that subsequent person-specific calibration modules like SCPT \cite{baoStoryYourEyes2021} can achieve even greater performance, highlighting the importance of a powerful, generalized feature extractor.
\section{\modelfull (\model) Architecture}
\label{sec:method}

\begin{figure*}[t]
    \centering
    \includegraphics[width=0.9\linewidth]{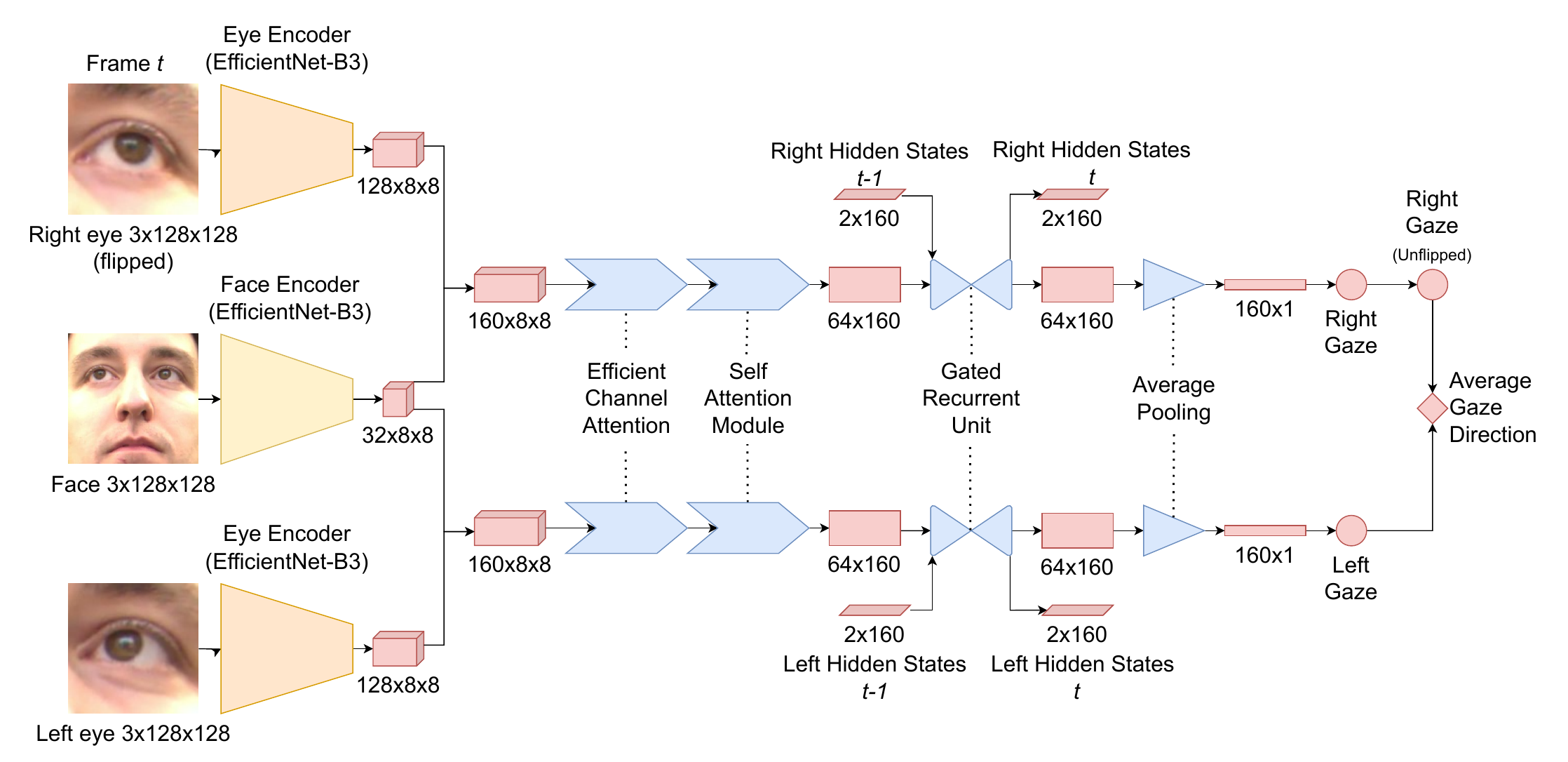}
    \caption{The architecture of our proposed \model{}. An input frame is processed by parallel eye and face encoders. The features are fused via ECA and a Self-Attention Module (SAM). A GRU first processes the feature maps spatially (intra-frame) before propagating its hidden state temporally (inter-frame) to the next frame. A final multi-layer perceptron (MLP) regresses the gaze vector. Input from EVE \cite{parkEndtoendVideobasedEyeTracking2020}.}
    \label{fig:model-arch}
\end{figure*}

Our proposed \textbf{\modelfull (\model)} is a video-based gaze estimation network designed to generate a robust spatio-temporal feature representation.
For a given frame, the model independently processes the left eye image with the face image, and the right eye image with the face image, yielding two separate gaze predictions.
These two predictions for the left and right eye are averaged to produce the final gaze vector.
For brevity, the following sections describe the architecture for a single eye-face stream.
The overall architecture of \model{} is depicted in Figure~\ref{fig:model-arch}

Our model consists of four key stages: 1) a multi-stream feature extraction backbone, 2) an attention-based feature fusion module, 3) our novel spatio-temporal recurrence, and 4) a final gaze regression head.
We detail each stage below.

\subsection{Multi-Stream Feature Extraction}
\label{subsec:extraction}

Building on the established success of hybrid CNN-Transformer architectures for gaze estimation \cite{chengGazeEstimationUsing2021}, the foundation of our model is a dual-stream backbone that processes eye and face patches independently, capturing their distinct characteristics.
While many prior works rely on ResNet variants \cite{heDeepResidualLearning2015, farkhondehSelfSupervisedGazeEstimation2022, liAppearanceBasedGazeEstimation2023a, jindalContrastiveRepresentationLearning2022, renGazeEstimationBased2023, parkEndtoendVideobasedEyeTracking2020, gideonUnsupervisedMultiViewGaze2022}, we employ a modified \textbf{EfficientNet-B3} \cite{tanEfficientNetRethinkingModel2020} as our feature extractor for both streams, selected for its superior performance in other computer vision tasks \cite{bahetiEffUNetNovelArchitecture2020, tanEfficientDetScalableEfficient2020, tanEfficientNetRethinkingModel2020} while still maintaining computational efficiency.
Specifically, we adapt the architecture by removing the final average pooling and prediction layers, and we modify the channel dimension in the final blocks to create two separate instances of the encoder:
\begin{itemize}
    \item \textbf{Eye Encoder}: This stream processes a $3 \times 128 \times 128$ eye patch. To maintain a consistent input geometry, images of the right eye are horizontally flipped before being passed to the encoder. The encoder is configured to output a feature map of size $128 \times 8 \times 8$.
    \item \textbf{Face Encoder}: This stream processes a $3 \times 128 \times 128$ face patch. The encoder is configured to output a feature map of size $32 \times 8 \times 8$.
\end{itemize}
Using separate encoders allows the model to learn specialized features for each modality.
For a given frame $I_t$, the encoders produce an eye feature map $\mathbf{E}_t \in \mathbb{R}^{128 \times 8 \times 8}$ and a face feature map $\mathbf{F}_t \in \mathbb{R}^{32 \times 8 \times 8}$.
Following Ren et al. \cite{renGazeEstimationBased2023}, we choose to represent the eye with more channels ($C_{\text{eye}}=128$) than the face ($C_{\text{face}}=32$), based on the established understanding that, while the information from the face helps guide the prediction \cite{baoAdaptiveFeatureFusion2021}, the eye is still the main contribution to the gaze direction.
These maps are then concatenated along the channel dimension to form a unified feature map $\mathbf{X}_t \in \mathbb{R}^{160 \times 8 \times 8}$, which serves as the input to the subsequent fusion stage.
For our ablation studies, we also implement a variant using ResNet \cite{heDeepResidualLearning2015} backbones, as detailed in our \textit{ResNetEncoder} implementation.

\subsection{Attention-Based Feature Fusion}
\label{sec:fusion}

Simply concatenating features from different domains can be suboptimal, as it may not effectively capture the complex relations between modalities \cite{baoAdaptiveFeatureFusion2021}. 
To fuse the eye and face features, we employ a two-stage attention mechanism as follows

\paragraph{Channel-wise Fusion.}
First, we apply an Efficient Channel Attention (ECA) module \cite{wangECANetEfficientChannel2020} to the concatenated feature map $\mathbf{X}_t$.
The ECA module adaptively re-weights the feature channels by learning their interdependencies without dimensionality reduction. This allows the model to dynamically emphasize either eye-specific details or broader facial context based on the input.
The kernel size of the 1D convolution within the ECA module is adaptively determined based on the channel dimension $C=160$, as described in \cite{wangECANetEfficientChannel2020}. 
The output is a channel-recalibrated feature map $\mathbf{X}'_t \in \mathbb{R}^{160 \times 8 \times 8}$.

\paragraph{Spatial Feature Learning.}
Next, we use a \textbf{Self-Attention Module (SAM)} \cite{vaswaniAttentionAllYou2017} to learn long-range spatial dependencies across the feature map. 
Inspired by the Vision Transformer (ViT) \cite{dosovitskiyImageWorth16x162021} and the work of Cheng and Lu \cite{chengGazeEstimationUsing2021} on integrating transformer with gaze estimation, we first reshape $\mathbf{X}'_t$ into a sequence of $S=64$ patches, where each patch is a feature vector of dimension $D=160$ and add a learnable positional encoding.
This sequence is then processed by a stack of $L=3$ Transformer encoder blocks with $8$ heads each. 
Each block consists of multi-head self-attention (MHSA) and a feed-forward network.
The output of the SAM is a sequence of spatially enriched feature vectors $\mathbf{Y}_t \in \mathbb{R}^{64 \times 160}$.

\subsection{Spatio-Temporal Recurrence}
\label{subsec:recurrence}

A limitation of prior temporal models for gaze \cite{liAppearanceBasedGazeEstimation2023a, parkEndtoendVideobasedEyeTracking2020, renGazeEstimationBased2023, zhouLearning3DGaze2019} is their reliance on premature spatial pooling. 
By applying spatial average pooling to the feature map, the entire spatially-aware feature map is collapsed into a single vector, losing the precise arrangement of facial features before the temporal dynamics are considered.

To overcome this, we propose a novel recurrence structure that models \textit{intra-frame} spatial context before modelling \textit{inter-frame} temporal dynamics.
We treat the output of our SAM, $\mathbf{Y}_t = \{y_1, y_2, \dots, y_{64}\}$, as an ordered \textbf{spatial sequence}, corresponding to a raster scan of the original $8\times8$ feature grid.
A Gated Recurrent Unit (GRU) \cite{choLearningPhraseRepresentations2014} with two layers is applied sequentially over these 64 patch vectors.
The initial hidden state for this spatial scan at time $t$, $h_{t,0}$, is the final hidden state from the \textit{previous time step's} spatial scan, $h_{t-1, 64}$.

Formally, the process is:
\begin{align}
    h_{0}^{(0)} &= 0^{2 \times 160} \\
    \mathbf{Z}^{(t)}, h_{64}^{(t)} &= \text{GRU}(\mathbf{Y}^{(t)}, h_{0}^{(t)}) \quad \text{for } t \ge 0 \label{eq:spatial_scan} \\
    h_{0}^{(t)} &= h_{64}^{(t-1)}  \quad \text{for } t \ge 1 \label{eq:temporal_prop}
\end{align}

Equation~\ref{eq:spatial_scan} describes the intra-frame scan, where the GRU iteratively processes the spatial patches within a single frame to produce a final, context-aware hidden state, $h_{64}^{(t)}$. Equation~\ref{eq:temporal_prop} describes the inter-frame propagation, where this summary state is passed forward in time to initialize the scan for the next frame.
This design allows the GRU to first build an understanding of the spatial relationships within a frame before this summary information is propagated through time.

\subsection{Gaze Regression and Loss Function}
\label{subsec:loss}

The sequence of spatially-aware features $\mathbf{Z}_t$ from the recurrent module is first aggregated through an average pooling layer across the spatial dimension.
This produces a single summary vector $\bar{\mathbf{z}}_t \in \mathbb{R}^{160}$ which is then passed to a two-layer MLP with a Tanh activation function to regress the 3D gaze direction vector $\mathbf{g}_t = (\text{pitch}, \text{yaw})$. 
The Tanh function constrains the output to $[-1,1]$, which we then scale by $\pi/2$ to map to the full range of gaze angles in radians.
The prediction for the right eye is finally horizontally flipped back.

\paragraph{Loss Function.}
Our model is trained end-to-end using a weighted combination of losses.
The primary loss is the angular error between the predicted gaze vector $\hat{\mathbf{g}}_t$ and the ground-truth vector $\mathbf{g}_t$.
\begin{equation}
    \mathcal{L}_{\text{ang}} = \arccos\left(\frac{\hat{\mathbf{g}}_t \cdot \mathbf{g}_t}{\|\hat{\mathbf{g}}_t\| \|\mathbf{g}_t\|}\right)
\end{equation}
Additionally, following Park et al. \cite{parkEndtoendVideobasedEyeTracking2020}, we incorporate losses on the predicted Point-of-Gaze (PoG) in both centimetre ($\mathcal{L}_{\text{PoG,cm}}$) and pixel ($\mathcal{L}_{\text{PoG,px}}$) coordinates.
The PoG is calculated from the gaze vector using the provided camera geometry, based on the implementation from the EVE framework. 
The total loss is:
\begin{equation}
    \mathcal{L}_{\text{total}} = \lambda_{\text{ang}} \mathcal{L}_{\text{ang}} + \lambda_{\text{px}} \mathcal{L}_{\text{PoG,px}} + \lambda_{\text{cm}} \mathcal{L}_{\text{PoG,cm}}
\end{equation}
where the $\lambda$ values are hyperparameters determined experimentally. 
Finally, we we adopt an offset augmentation strategy during training with a standard deviation of $3^\circ$, following the protocol of Park et al. \cite{parkEndtoendVideobasedEyeTracking2020}.
\section{Experiments}
\label{sec:expe}

In this section, we present a comprehensive evaluation of our \model.
We first detail our experimental setup, including the dataset, evaluation metrics, and implementation details.
We then compare \model{} against state-of-the-art methods on the EVE benchmark \cite{parkEndtoendVideobasedEyeTracking2020}.
Finally, we conduct a series of ablation studies to disentangle the contributions of each component of our architecture and validate our design choices.

\subsection{Setup}
\label{subsec:setup}

\paragraph{Dataset.}
We conduct all experiments on the large-scale EVE dataset \cite{parkEndtoendVideobasedEyeTracking2020}. 
We selected EVE as our primary benchmark, as its focus on continuous video of natural user-screen interactions is ideal for evaluating temporal gaze models in desktop and laptop scenarios. 
The dataset contains over 12 million frames from 54 participants, featuring diverse eye movements and significant head pose variation. 
Data was collected across four camera views (basler and three webcams at top-left, -centre, and -right) while participants viewed various stimuli (images, videos, Wikipedia pages). 
We follow the official protocol from Park et al., training on the designated training split. 
Final performance and key ablations are reported on the test set, with additional detailed results on the validation set available in the supplementary material.
    
\paragraph{Evaluation Metrics.}
Following prior work \cite{parkEndtoendVideobasedEyeTracking2020, baoStoryYourEyes2021}, we use the \textbf{Angular Error} as our primary evaluation metric, measured in degrees ($^\circ$) between the predicted 3D gaze vector and the ground-truth vector. 
As a complementary metric, we also report the Point-of-Gaze (PoG) Error, which is the Euclidean distance in centimetres (cm) between the predicted and actual gaze points on the screen.
For fair comparison, all results are reported on the final test set, accessible via the Codalab competition \cite{codalab_competitions_JMLR} set up by the authors of the EVE dataset.\footnote{Competition URL: \url{https://codalab.lisn.upsaclay.fr/competitions/11397}}
The test set labels are held out, and final performance is evaluated by submitting predictions to the official Codalab competition server.

\paragraph{Implementation Details.}
Our \model{} is implemented in Pytorch using an EfficientNet-B3 backbone, pre-trained on ImageNet \cite{tanEfficientNetRethinkingModel2020}.
The model is trained end-to-end for 4 epochs using the Adam optimizer \cite{kingmaAdamMethodStochastic2017} with a batch size of 6 and a base learning rate of $1e^{-4}$, which is scaled linearly with the batch size and follows a cosine annealing schedule.
Our total loss is a weighted sum of angular error ($\lambda_{ang}=1.0$) and PoG error in cm ($\lambda_{cm}=1e^{-2}$) determined empirically.
For experiments with the SCPT modules \cite{baoStoryYourEyes2021}, we use our trained \model{} as a frozen feature extractor and use the publicly available SCPT weights for evaluation.

\paragraph{Model Complexity.}
We analyse the computational cost and parameter distribution of \model{} to contextualize its performance. 
For a single-frame, dual-stream forward pass on the input described in Section~\ref{subsec:extraction}, our model computes $6.39$ GFLOPs.
Additionally, on a PC with a \textit{Nvidia RTX 4090} and an \textit{Intel Core Ultra 9 285K}, it achieves an inference speed of \textbf{$105$ FPS}, using $800$ MB of VRAM.
Finally, the vast majority of the \textbf{21 M} parameters in our model are concentrated in the dual EfficientNet backbones used for feature extraction (over $94\%$). 
In contrast, our proposed combination of attention and spatio-temporal recurrence modules is highly efficient and comprise less than $6\%$ of the total parameters.
A detailed breakdown of the parameters' distribution can be found in the supplementary material.

\subsection{Comparison with State-of-the-Art}
\label{subsec:sota}
 
\begin{table}[hbt]
    \centering
    \caption{Comparison with state-of-the-art methods on the EVE \textit{test} set. All methods listed are appearance-based, using single-view camera input. \textbf{Bold} indicates the best performance. *Results reported by \cite{renGazeEstimationBased2023}.}
    \label{tab:sota_results}
    \begin{tabular}{lcc}
        \toprule
            \textbf{Method} & \shortstack{\textbf{Angular}\\\textbf{Error ($^{\circ}$)}} $\downarrow$ & \shortstack{\textbf{Euclidean}\\\textbf{PoG Error (cm)}} $\downarrow$ \\
        \midrule
            Hybrid-ViT \cite{chengGazeEstimationUsing2021}* & 3.54 & 3.92 \\
            EyeNet-GRU \cite{parkEndtoendVideobasedEyeTracking2020} & 3.48 & 3.85 \\
            Gaze360 \cite{kellnhoferGaze360PhysicallyUnconstrained2019}* & 3.45 & 3.83 \\
            FE-NET \cite{renGazeEstimationBased2023} & 3.19 & 3.55 \\
        \midrule
            \textbf{\model{} (Ours)} & \textbf{2.58} & \textbf{2.87} \\
        \bottomrule
    \end{tabular}
\end{table}

We first evaluate the performance of \model{} as a standalone gaze estimator against several recent and influential methods in video-based gaze estimation.
We select baselines that represent different architectural approaches, including the original EVE benchmark model (EyeNet-GRU \cite{parkEndtoendVideobasedEyeTracking2020}), Transformer-based (Hybrid-ViT \cite{chengGazeEstimationUsing2021}), a similar workflow to ours (FE-NET \cite{renGazeEstimationBased2023}), and other state-of-the-art architectures (Gaze360 \cite{kellnhoferGaze360PhysicallyUnconstrained2019}).
All baseline results are taken from their original publications or subsequent papers that use the official EVE evaluation protocol, ensuring a fair comparison.
As shown in Table~\ref{tab:sota_results}, \model{} achieves a new state-of-the-art performance on the test set.
Our model achieves an angular error of \textbf{$2.58^{\circ}$}, a $25.9\%$ relative improvement over the original EyeNet-GRU baseline \cite{parkEndtoendVideobasedEyeTracking2020}, and a $19.1\%$ relative improvement over FE-NET \cite{renGazeEstimationBased2023}, the best-performing, single-view model on EVE.
This result validates that our architecture's focus on superior feature representation translates into substantial performance gains.
Moreover, our method achieves comparable performance to other 
methods that leverage \textit{additional} input modalities. GazeRefineNet \cite{parkEndtoendVideobasedEyeTracking2020} and DVGaze \cite{chengDVGazeDualViewGaze2023} both reported an error of $2.49^\circ$, but require either the screen content, which may raise privacy concerns for continuous screen capture, or a dual-camera setup, respectively.
Our method uses only a single camera view, making it more broadly applicable.

\subsection{Performance as a Backbone for Person-Specific Adaptation}
\label{subsec:scpt}

\begin{table}[h]
    \centering
    \caption{Performance comparison on the EVE \textit{test} set when used as a backbone for the SCPT adaptation modules~\cite{baoStoryYourEyes2021}. Our model provides a superior initial prediction, leading to a new overall SOTA result.}
    \label{tab:scpt_results}
    \begin{tabular}{lcc}
        \toprule
            \textbf{Backbone + SCPT Modules} & \multicolumn{2}{c}{\textbf{Angular Error ($^\circ$) $\downarrow$}} \\
        \cmidrule(lr){2-3}
        & \textbf{Offline} & \textbf{Online} \\
        \midrule
            EyeNet~\cite{parkEndtoendVideobasedEyeTracking2020} + SCPT & 2.15 & 1.95 \\
            \textbf{\model{} (Ours) + SCPT} & \textbf{2.03} & \textbf{1.87} \\
            \quad \textit{w/o ECA Module} & 2.15 & 1.97 \\
        \bottomrule
    \end{tabular}
\end{table}

To demonstrate the quality of our learned feature representation, we further evaluate its effectiveness as a backbone for person-specific adaptation methods.
We use our trained \model{} to provide initial gaze predictions for the Self-Calibration and Person-specific Transform (SCPT) modules from Bao et al.~\cite{baoStoryYourEyes2021}.
SCPT is a state-of-the-art unsupervised adaptation technique that refines gaze predictions for a specific person by modelling their unique offsets from a collection of their gaze data, without requiring ground-truth labels. 
The adaptation can be performed \textbf{online}, using only past frames in a sequence, or \textbf{offline}, using all available frames for a person to achieve maximum accuracy.
As shown in Table~\ref{tab:scpt_results}, our model provides a superior initial prediction, which in turn boosts the performance of SCPT in both online and offline settings
We compare against the original EyeNet backbone, which was the baseline used in the SCPT paper, providing a direct and fair comparison.
Using \model{} as the feature extractor reduces the error from $2.15^\circ$ (online) and $1.95^\circ$ (offline) to \textbf{$2.03^\circ$} (online) and \textbf{$1.87^\circ$} (offline).
These results underscore that a stronger generalized feature extractor is crucial for maximizing the potential of downstream personalization techniques.

\subsection{Ablation Study}
\label{subsec:ablation}

To validate our architectural design and understand the contribution of each component, we conduct a comprehensive ablation study.
We start with our full \model{} and systematically remove or alter key modules.
All ablation experiments are evaluated on the EVE validation and test set.

\begin{table}[h]
    \centering
    \caption{Ablation study of the core modules in \model{} on the EVE \textit{test} set. The significant performance drop upon removing the SAM and the GRU, and particularly the degradation from altering the recurrence structure, validates our key design choices. \textbf{Bold} indicates the best performance.}
    \label{tab:ablation_modules}
    \begin{tabular}{lcc}
        \toprule
            \textbf{Model Configuration} & \multicolumn{2}{c}{\textbf{Error $\downarrow$}} \\
            \cmidrule(lr){2-3} & \textbf{Angular ($^{\circ}$)} & \textbf{PoG (cm)} \\
        \midrule
            \textbf{\model{} (Full Model)} & 2.58 & 2.87 \\
        \midrule
            \textit{Ablating Fusion Modules:} \\
            \quad w/o ECA Module & \textbf{2.56} & \textbf{2.86} \\
            \quad w/o SAM & 4.84 & 5.36 \\
        \midrule
            \textit{Ablating Recurrence:} \\
            \quad w/o GRU & 2.88 & 3.21 \\
            \quad Spatial Pooling pre-GRU & 2.79 & 3.13 \\
        \bottomrule
    \end{tabular}
\end{table}

\paragraph{Impact of Core Architectural Components.}
Table~\ref{tab:ablation_modules} presents the results of ablating our feature fusion and temporal modelling components.
Our full model serves as the baseline, achieving a $2.58^\circ$ angular error and $2.87$ cm PoG error on the test set.
The impact of removing the SAM is dramatic, with the angular error (resp. PoG error) increasing to $4.84^\circ$ (resp. $5.36$ cm), highlighting its critical role in learning generalizable spatial relationships.
The effect of the ECA module is more nuanced. 
While its removal results in a marginal $0.02^\circ$ (resp. $0.01$ cm) improvement on the angular error (resp. PoG error), its contribution becomes evident when evaluating the quality of the learned features for downstream adaptation. 
As shown in Table~\ref{tab:scpt_results}, the features learned \textit{without} ECA are less effective for person-specific adaptation, resulting in a higher final error with the SCPT modules ($1.97^\circ$ vs. $1.87^\circ$ for the angular error). 
This demonstrates that ECA helps produce a more robust feature representation, aligning with our work's primary goal, even if it does not improve raw generalization in this instance.

Disabling the temporal GRU module entirely increases the angular error to $2.88^\circ$ and PoG error to $3.21$ cm, confirming the importance of temporal modelling.
Crucially, when we revert to the conventional approach of spatially pooling features \textit{before} the GRU, the angular error increases to $2.79^\circ$ and the PoG error to $3.10$ cm.
These results validate our hypothesis: preserving and modelling intra-frame spatial context with our spatio-temporal recurrence leads to lower gaze estimation error than premature spatial pooling.

\begin{table}[h]
    \centering
    \caption{Ablation study on the backbone architecture and fusion strategy on the EVE \textit{test} set. \textbf{Bold} indicates the best performance.}
    \label{tab:ablation_backbone}
    \begin{tabular}{lcc}
        \toprule
            \textbf{Backbone Configuration} &  \multicolumn{2}{c}{\textbf{Error $\downarrow$}} \\
            \cmidrule(lr){2-3} & \textbf{Angular ($^{\circ}$)} & \textbf{PoG (cm)} \\
        \midrule
            \textbf{EfficientNet-B3 (\model)} & \textbf{2.58} & \textbf{2.87} \\
        \midrule
            ResNet-18 & 4.22 & 4.99 \\
            EfficientNet-B0 & 2.78 & 3.09 \\
            EfficientNet-B7 & 2.90 & 3.24 \\
            EfficientNet-V2-S & 3.18 & 3.55 \\
        \midrule
            Early Fusion Strategy & 3.26 & 3.66 \\
        \bottomrule
    \end{tabular}
\end{table}

\paragraph{Impact of Backbone and Fusion Strategy.}
To validate our specific architectural choices, we conduct further experiments analysing the impact of the backbone and the feature fusion strategy.
The results on the test set, presented in Table~\ref{tab:ablation_backbone}, highlight the importance of these decisions for model generalization.
First, we replace the EfficientNet-B3 backbone with several alternatives while keeping the rest of the \model{} architecture constant.
With a ResNet-18 \cite{heDeepResidualLearning2015} backbone, the performance collapses to $4.22^\circ$, indicating the failure to generalise.
Within the EfficientNet family, our choice of B3 proves to be optimal.
A simpler EfficientNet-B0 yields a higher angular error of $2.78^\circ$, while a larger and more complex EfficientNet-B7 also sees performance degrade to $2.90^\circ$.
Similarly, a more recent EfficientNet-V2-S \cite{tanEfficientNetV2SmallerModels2021} backbone results in a $3.18^\circ$ angular error. 
This suggests that simply increasing model capacity does not guarantee better performance, and that EfficientNet-B3 offers the best feature representation for this task, and our fusion and recurrence modules.

Second, we investigate an early fusion strategy where features from the left eye, right eye, and face are combined before being processed by a single encoder stream.
This approach performs poorly, with an angular error of $3.26^\circ$.
Overall, these ablations confirm that our choice of an EfficientNet-B3 backbone and a late-fusion strategy are critical contributors to the strong generalisation performance of \model{}.

\subsection{Qualitative Evaluation}
\label{subsec:qualitative_eval}

Beyond aggregate metrics, we analyse our model's performance across different conditions and participants to understand its robustness and limitations. 
Due to the test labels and detailed results being held out, the figures presented for the test performances rely on the mean angular errors and standard deviations provided by the Codalab platform.
The detailed breakdown of our results on the validation set can be found in the supplementary material.

\begin{figure}[h]
    \centering
    \begin{subfigure}[b]{0.98\linewidth}
        \centering
        \includegraphics[width=\textwidth]{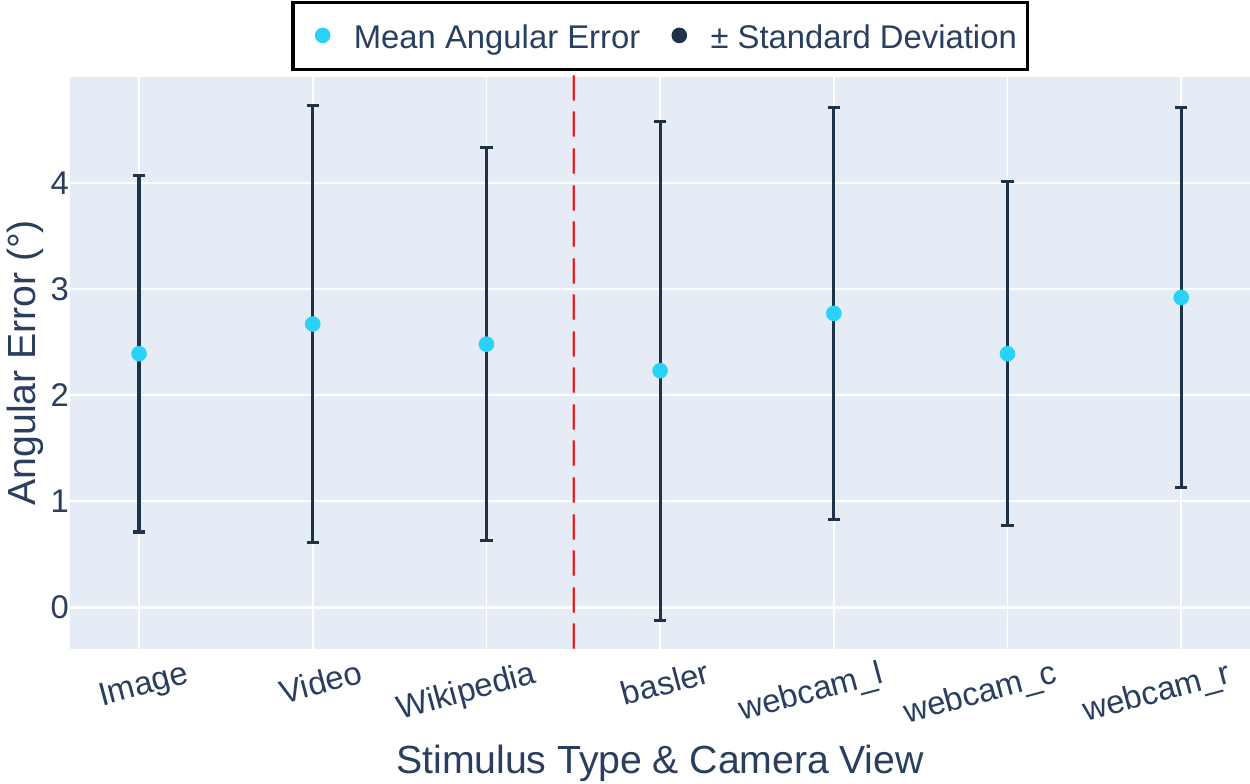}
        \caption{Mean angular error and standard deviation across different stimuli (image, video, and wikipedia) and camera views.}
        \label{fig:test_error_stim_cam}
    \end{subfigure}
    
    \vspace{0.5cm}
    
    \begin{subfigure}[b]{0.98\linewidth}
        \centering
        \includegraphics[width=\textwidth]{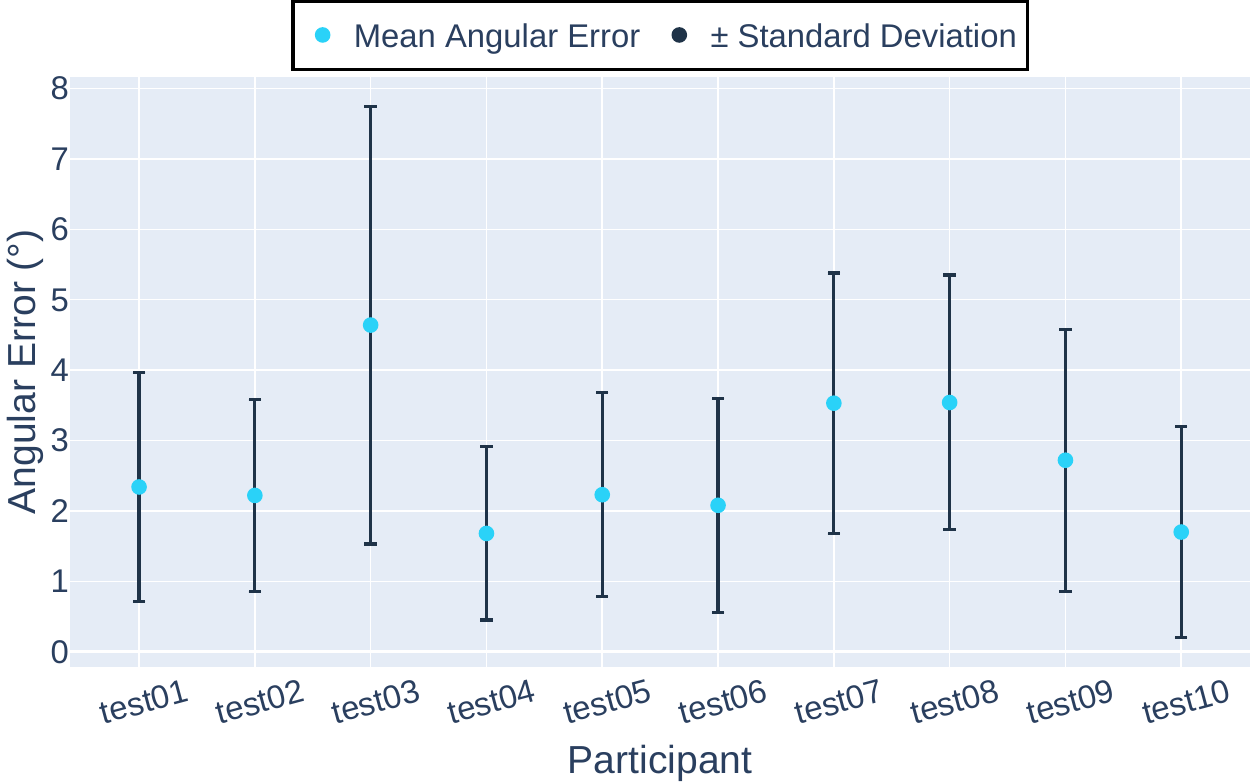}
        \caption{Mean angular error and standard deviation across individual participants.}
        \label{fig:test_part}
    \end{subfigure}
    
    \caption{Analysis of \model{}'s robustness on the EVE \textbf{test set}. (a) Performance is stable across stimuli and camera views, with the centre webcam (\textit{webcam\_c}) showing the lowest variance. (b) Performance varies significantly across participants, highlighting the challenge of person-specific factors.}
    \label{fig:qualitative_analysis}
\end{figure}

Figure~\ref{fig:qualitative_analysis} presents this breakdown. Figure~\ref{fig:test_error_stim_cam} analyses performance across different camera views and stimulus types.
Performance is largely consistent across the three stimuli, with \textit{Video} content yielding a slightly higher error, likely due to the more complex and rapid eye movements it elicits compared to static \textit{Image} and \textit{Wikipedia} pages.
A more pronounced trend is observed across camera views. 
While all views perform well, the centrally-mounted \textit{webcam\_c} not only achieves low error but also exhibits the lowest variance. 
This is a promising result for real-world usability, as the \textit{camera\_c} position corresponds to where a webcam would typically be placed on a laptop or monitor. 

Figure~\ref{fig:test_part} reveals the significant impact of person-specific factors, which remains a central challenge in generalized gaze estimation. The test participants can be broadly clustered into distinct performance groups. A majority of participants (e.g., \textit{test01, test02, test04}) form a high-performance cluster, with mean errors around $2.0^\circ$. Another group (e.g., \textit{test07, test08}) shows moderate performance with errors around $3.0^\circ$. Crucially, one participant, \textit{test03}, is a clear outlier, exhibiting a much higher mean error of $4.6^\circ$. 
It is precisely for these outlier cases that downstream person-specific adaptation methods like SCPT are essential, as they are designed to learn and compensate for these unique, individual-specific characteristics.
Our analysis of the SCPT results confirms this, but with a nuance.
For the high-performance cluster, SCPT provides a moderate but consistent error reduction of $20\%$ to $30\%$.
Its impact is most dramatic on the mid-tier performers, such as \textit{test07}, where it reduces the error by $58\%$.
However, for the most challenging outlier, \textit{test03}, the improvement is minimal, at less than $10\%$.
This suggests a key distinction: SCPT is highly effective at correcting for person-specific geometric and anatomical biases, but its effectiveness is limited when the primary source of error stems from persistent issues in the input data. 
As we will illustrate next, the high error for participant \textit{test03} appears to be caused by such an input challenge.

\begin{figure}[h]
    \centering
    \includegraphics[width=\linewidth]{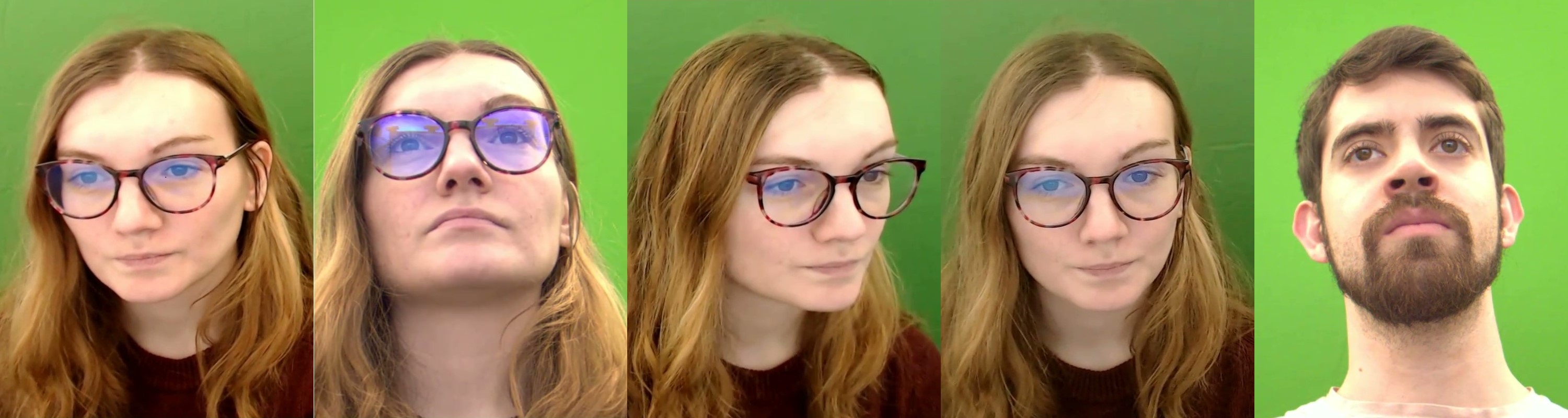}
    \caption{Qualitative examples from the EVE test set \cite{parkEndtoendVideobasedEyeTracking2020}. The four leftmost frames show a challenging case (Participant \textit{test03}) while the rightmost frame shows a high-performance case (Participant \textit{test04}). \textit{Images reproduced with permission from authors.}}
    \label{fig:qualitative_examples}
\end{figure}

To visually investigate these performance differences, Figure~\ref{fig:qualitative_examples} presents example frames from two representative participants.
The four leftmost frames show participant \textit{test03}, the most challenging case in the test set. 
In each view, a strong blue light reflection on the participant's eyeglasses occludes the eye region. 
This persistent visual artifact severely degrades the quality of the input features, naturally leading to a high mean prediction error of $4.64^\circ$ and a standard deviation of $3.11^\circ$ as reported on Figure~\ref{fig:test_part}. 
In contrast, the rightmost frame shows participant \textit{test04}, one of the best-performing participants with a mean prediction error of $1.68^\circ$ and a standard deviation of $1.23^\circ$. 
The eye region is clearly visible with even lighting, and the pupil's position is unambiguous, allowing our model to extract high-quality features and make an accurate prediction. 
These examples visually confirm that performance remains heavily dependent on input image quality, with factors like eyeglass reflections being a significant hurdle for all appearance-based methods.
\section{Conclusion}
\label{sec:conclusion}

In this paper, we addressed the critical challenge of feature representation in video-based gaze estimation. We proposed the \modelfull{} (\model{}), a novel architecture designed to overcome the limitations of premature spatial pooling found in prior temporal models. 
Our approach integrates the EfficientNet-B3 backbone with a two-stage attention mechanism for effective eye-face feature fusion. 
A key contribution is our spatio-temporal recurrence, which first models intra-frame spatial context by treating the feature map as a sequence, before propagating this contextual summary through time.

Our experiments on the challenging EVE dataset demonstrate the effectiveness of this design. 
\model{} achieves a new state-of-the-art angular error of $2.58^\circ$ for single-view 3D gaze estimation. 
Furthermore, we showed that the superior feature representation learned by our model serves as a powerful backbone for downstream person-specific adaptation, boosting the performance with the SCPT modules to a final error of $1.87^\circ$. 
Our comprehensive ablation study validated our architectural choices, confirming that the self-attention module and our novel recurrence structure are key drivers of the model's strong generalization performance.

The robust spatio-temporal features learned by \model{} open up several compelling avenues for future research. One promising direction is moving beyond gaze direction to predict higher-level cognitive states, such as cognitive load, confusion, or attention, where the subtle temporal dynamics captured by our model could be beneficial.
Furthermore, as our qualitative analysis highlights, a persistent challenge for all appearance-based methods is robustness to visual artifacts like eyeglass reflections. 
Future work could explicitly tackle this by developing noise-aware training strategies or generative models to reconstruct occluded eye regions, further improving real-world applicability.
\section{Acknowledgements}
\label{sec:acknowledgements}

The computing resources and services used in this work were provided by the VSC (Flemish Supercomputer Center), funded by the Research Foundation - Flanders (FWO) and the Flemish Government.
This research is supported by Internal Funds KU Leuven (STG/23/054).
{
    \small
    \bibliographystyle{ieeenat_fullname}
    \bibliography{main}

\begin{thebibliography}{47}
\providecommand{\natexlab}[1]{#1}
\providecommand{\url}[1]{\texttt{#1}}
\expandafter\ifx\csname urlstyle\endcsname\relax
  \providecommand{\doi}[1]{doi: #1}\else
  \providecommand{\doi}{doi: \begingroup \urlstyle{rm}\Url}\fi

\bibitem[Akinyelu and
  Blignaut(2020)]{akinyeluConvolutionalNeuralNetworkBased2020}
Andronicus~A. Akinyelu and Pieter Blignaut.
\newblock Convolutional {{Neural Network-Based Methods}} for {{Eye Gaze
  Estimation}}: {{A Survey}}.
\newblock \emph{IEEE Access}, 8:\penalty0 142581--142605, 2020.

\bibitem[Baheti et~al.(2020)Baheti, Innani, Gajre, and
  Talbar]{bahetiEffUNetNovelArchitecture2020}
Bhakti Baheti, Shubham Innani, Suhas Gajre, and Sanjay Talbar.
\newblock Eff-{{UNet}}: {{A Novel Architecture}} for {{Semantic Segmentation}}
  in {{Unstructured Environment}}.
\newblock In \emph{2020 {{IEEE}}/{{CVF Conference}} on {{Computer Vision}} and
  {{Pattern Recognition Workshops}} ({{CVPRW}})}, pages 1473--1481, 2020.

\bibitem[Ballas et~al.(2016)Ballas, Yao, Pal, and
  Courville]{ballasDelvingDeeperConvolutional2016}
Nicolas Ballas, Li Yao, Chris Pal, and Aaron Courville.
\newblock Delving {{Deeper}} into {{Convolutional Networks}} for {{Learning
  Video Representations}}, 2016.

\bibitem[Bao et~al.(2021{\natexlab{a}})Bao, Liu, and Yu]{baoStoryYourEyes2021}
Jun Bao, Buyu Liu, and Jun Yu.
\newblock The {{Story}} in {{Your Eyes}}: {{An Individual-difference-aware
  Model}} for {{Cross-person Gaze Estimation}}.
\newblock https://arxiv.org/abs/2106.14183v1, 2021{\natexlab{a}}.

\bibitem[Bao et~al.(2021{\natexlab{b}})Bao, Cheng, Liu, and
  Lu]{baoAdaptiveFeatureFusion2021}
Yiwei Bao, Yihua Cheng, Yunfei Liu, and Feng Lu.
\newblock Adaptive {{Feature Fusion Network}} for {{Gaze Tracking}} in {{Mobile
  Tablets}}, 2021{\natexlab{b}}.

\bibitem[Cheng and Lu(2021)]{chengGazeEstimationUsing2021}
Yihua Cheng and Feng Lu.
\newblock Gaze {{Estimation}} using {{Transformer}}, 2021.

\bibitem[Cheng and Lu(2023)]{chengDVGazeDualViewGaze2023}
Yihua Cheng and Feng Lu.
\newblock {{DVGaze}}: {{Dual-View Gaze Estimation}}.
\newblock In \emph{2023 {{IEEE}}/{{CVF International Conference}} on {{Computer
  Vision}} ({{ICCV}})}, pages 20575--20584, Paris, France, 2023. IEEE.

\bibitem[Cheng et~al.(2024)Cheng, Wang, Bao, and
  Lu]{chengAppearancebasedGazeEstimation2024}
Yihua Cheng, Haofei Wang, Yiwei Bao, and Feng Lu.
\newblock Appearance-based {{Gaze Estimation}} with {{Deep Learning}}: {{A
  Review}} and {{Benchmark}}.
\newblock \emph{IEEE Transactions on Pattern Analysis and Machine
  Intelligence}, pages 1--20, 2024.

\bibitem[Cho et~al.(2014)Cho, van Merrienboer, Gulcehre, Bahdanau, Bougares,
  Schwenk, and Bengio]{choLearningPhraseRepresentations2014}
Kyunghyun Cho, Bart van Merrienboer, Caglar Gulcehre, Dzmitry Bahdanau, Fethi
  Bougares, Holger Schwenk, and Yoshua Bengio.
\newblock Learning {{Phrase Representations}} using {{RNN Encoder-Decoder}} for
  {{Statistical Machine Translation}}, 2014.

\bibitem[Dosovitskiy et~al.(2021)Dosovitskiy, Beyer, Kolesnikov, Weissenborn,
  Zhai, Unterthiner, Dehghani, Minderer, Heigold, Gelly, Uszkoreit, and
  Houlsby]{dosovitskiyImageWorth16x162021}
Alexey Dosovitskiy, Lucas Beyer, Alexander Kolesnikov, Dirk Weissenborn,
  Xiaohua Zhai, Thomas Unterthiner, Mostafa Dehghani, Matthias Minderer, Georg
  Heigold, Sylvain Gelly, Jakob Uszkoreit, and Neil Houlsby.
\newblock An {{Image}} is {{Worth}} 16x16 {{Words}}: {{Transformers}} for
  {{Image Recognition}} at {{Scale}}, 2021.

\bibitem[Farkhondeh et~al.(2022)Farkhondeh, Palmero, Scardapane, and
  Escalera]{farkhondehSelfSupervisedGazeEstimation2022}
Arya Farkhondeh, Cristina Palmero, Simone Scardapane, and Sergio Escalera.
\newblock Towards {{Self-Supervised Gaze Estimation}}, 2022.

\bibitem[Gideon et~al.(2022)Gideon, Su, and
  Stent]{gideonUnsupervisedMultiViewGaze2022}
John Gideon, Shan Su, and Simon Stent.
\newblock Unsupervised {{Multi-View Gaze Representation Learning}}.
\newblock In \emph{2022 {{IEEE}}/{{CVF Conference}} on {{Computer Vision}} and
  {{Pattern Recognition Workshops}} ({{CVPRW}})}, pages 4997--5005, New
  Orleans, LA, USA, 2022. IEEE.

\bibitem[Guestrin and Eizenman(2006)]{guestrinGeneralTheoryRemote2006}
E.D. Guestrin and M. Eizenman.
\newblock General theory of remote gaze estimation using the pupil center and
  corneal reflections.
\newblock \emph{IEEE Transactions on Biomedical Engineering}, 53\penalty0
  (6):\penalty0 1124--1133, 2006.

\bibitem[Hansen and {Qiang Ji}(2010)]{hansenEyeBeholderSurvey2010}
D.W. Hansen and {Qiang Ji}.
\newblock In the {{Eye}} of the {{Beholder}}: {{A Survey}} of {{Models}} for
  {{Eyes}} and {{Gaze}}.
\newblock \emph{IEEE Transactions on Pattern Analysis and Machine
  Intelligence}, 32\penalty0 (3):\penalty0 478--500, 2010.

\bibitem[Harezlak and Kasprowski(2018)]{harezlakApplicationEyeTracking2018}
Katarzyna Harezlak and Pawel Kasprowski.
\newblock Application of eye tracking in medicine: {{A}} survey, research
  issues and challenges.
\newblock \emph{Computerized Medical Imaging and Graphics}, 65:\penalty0
  176--190, 2018.

\bibitem[He et~al.(2015)He, Zhang, Ren, and Sun]{heDeepResidualLearning2015}
Kaiming He, Xiangyu Zhang, Shaoqing Ren, and Jian Sun.
\newblock Deep {{Residual Learning}} for {{Image Recognition}}, 2015.

\bibitem[He et~al.(2024)He, Xie, Zhang, and
  Dai]{heIntegratingEyeTrackingGames2024}
Yu He, Shiwei Xie, Han Zhang, and Yang~Guanghui Dai.
\newblock Integrating {{Eye-Tracking}} in {{Games}}: {{Enhancing Player
  Experience}} and {{Understanding Social Gaze Issues}}.
\newblock In \emph{2024 {{IEEE}} 13th {{Global Conference}} on {{Consumer
  Electronics}} ({{GCCE}})}, pages 1396--1400, 2024.

\bibitem[Jin et~al.(2023)Jin, Dai, and Nguyen]{jinKappaAngleRegression2023}
Shiwei Jin, Ji Dai, and Truong Nguyen.
\newblock Kappa {{Angle Regression}} with {{Ocular Counter-Rolling Awareness}}
  for {{Gaze Estimation}}.
\newblock In \emph{2023 {{IEEE}}/{{CVF Conference}} on {{Computer Vision}} and
  {{Pattern Recognition Workshops}} ({{CVPRW}})}, pages 2659--2668, Vancouver,
  BC, Canada, 2023. IEEE.

\bibitem[Jindal and
  Manduchi(2022)]{jindalContrastiveRepresentationLearning2022}
Swati Jindal and Roberto Manduchi.
\newblock Contrastive {{Representation Learning}} for {{Gaze Estimation}}.
\newblock https://arxiv.org/abs/2210.13404v1, 2022.

\bibitem[Kellnhofer et~al.(2019)Kellnhofer, Recasens, Stent, Matusik, and
  Torralba]{kellnhoferGaze360PhysicallyUnconstrained2019}
Petr Kellnhofer, Adria Recasens, Simon Stent, Wojciech Matusik, and Antonio
  Torralba.
\newblock Gaze360: {{Physically Unconstrained Gaze Estimation}} in the
  {{Wild}}, 2019.

\bibitem[Kingma and Ba(2017)]{kingmaAdamMethodStochastic2017}
Diederik~P. Kingma and Jimmy Ba.
\newblock Adam: {{A Method}} for {{Stochastic Optimization}}, 2017.

\bibitem[Li et~al.(2023)Li, Huang, Chen, Wang, and
  Tan]{liAppearanceBasedGazeEstimation2023a}
Yujie Li, Longzhao Huang, Jiahui Chen, Xiwen Wang, and Benying Tan.
\newblock Appearance-{{Based Gaze Estimation Method Using Static Transformer
  Temporal Differential Network}}.
\newblock \emph{Mathematics}, 11\penalty0 (3):\penalty0 686, 2023.

\bibitem[Lind{\'e}n et~al.(2019)Lind{\'e}n, Sj{\"o}strand, and
  Proutiere]{lindenLearningPersonalizeAppearanceBased2019}
Erik Lind{\'e}n, Jonas Sj{\"o}strand, and Alexandre Proutiere.
\newblock Learning to {{Personalize}} in {{Appearance-Based Gaze Tracking}},
  2019.

\bibitem[Liu et~al.(2019)Liu, Wu, Sun, and
  Lee]{liuSpatioTemporalGRUTrajectory2019}
Hongbin Liu, Hao Wu, Weiwei Sun, and Ickjai Lee.
\newblock Spatio-{{Temporal GRU}} for {{Trajectory Classification}}.
\newblock In \emph{2019 {{IEEE International Conference}} on {{Data Mining}}
  ({{ICDM}})}, pages 1228--1233, 2019.

\bibitem[Majaranta and Bulling(2014)]{majarantaEyeTrackingEyeBased2014}
P{\"a}ivi Majaranta and Andreas Bulling.
\newblock Eye {{Tracking}} and {{Eye-Based Human}}--{{Computer Interaction}}.
\newblock In \emph{Advances in {{Physiological Computing}}}, pages 39--65.
  Springer, London, 2014.

\bibitem[Nagpure and Okuma(2023)]{nagpureSearchingEfficientNeural2023}
Vikrant Nagpure and Kenji Okuma.
\newblock Searching {{Efficient Neural Architecture With Multi-Resolution
  Fusion Transformer}} for {{Appearance-Based Gaze Estimation}}.
\newblock In \emph{Proceedings of the {{IEEE}}/{{CVF Winter Conference}} on
  {{Applications}} of {{Computer Vision}}}, pages 890--899, 2023.

\bibitem[Park et~al.(2019)Park, Mello, Molchanov, Iqbal, Hilliges, and
  Kautz]{parkFewShotAdaptiveGaze2019a}
Seonwook Park, Shalini~De Mello, Pavlo Molchanov, Umar Iqbal, Otmar Hilliges,
  and Jan Kautz.
\newblock Few-{{Shot Adaptive Gaze Estimation}}.
\newblock In \emph{2019 {{IEEE}}/{{CVF International Conference}} on {{Computer
  Vision}} ({{ICCV}})}, pages 9367--9376, Seoul, Korea (South), 2019. IEEE.

\bibitem[Park et~al.(2020)Park, Aksan, Zhang, and
  Hilliges]{parkEndtoendVideobasedEyeTracking2020}
Seonwook Park, Emre Aksan, Xucong Zhang, and Otmar Hilliges.
\newblock Towards {{End-to-end Video-based Eye-Tracking}}.
\newblock https://arxiv.org/abs/2007.13120v1, 2020.

\bibitem[Pathirana et~al.(2022)Pathirana, Senarath, Meedeniya, and
  Jayarathna]{pathiranaEyeGazeEstimation2022}
Primesh Pathirana, Shashimal Senarath, Dulani Meedeniya, and Sampath
  Jayarathna.
\newblock Eye gaze estimation: {{A}} survey on deep learning-based approaches.
\newblock \emph{Expert Systems with Applications}, 199:\penalty0 116894, 2022.

\bibitem[Pavao et~al.(2023)Pavao, Guyon, Letournel, Tran, Baro, Escalante,
  Escalera, Thomas, and Xu]{codalab_competitions_JMLR}
Adrien Pavao, Isabelle Guyon, Anne-Catherine Letournel, Dinh-Tuan Tran, Xavier
  Baro, Hugo~Jair Escalante, Sergio Escalera, Tyler Thomas, and Zhen Xu.
\newblock Codalab competitions: An open source platform to organize scientific
  challenges.
\newblock \emph{Journal of Machine Learning Research}, 24\penalty0
  (198):\penalty0 1--6, 2023.

\bibitem[Pereira and Hussain(2024)]{pereiraReviewTransformerBasedModels2024}
Gracile~Astlin Pereira and Muhammad Hussain.
\newblock A {{Review}} of {{Transformer-Based Models}} for {{Computer Vision
  Tasks}}: {{Capturing Global Context}} and {{Spatial Relationships}}, 2024.

\bibitem[Peters et~al.(2005)Peters, Pelachaud, Bevacqua, Mancini, and
  Poggi]{petersModelAttentionInterest2005}
Christopher Peters, Catherine Pelachaud, Elisabetta Bevacqua, Maurizio Mancini,
  and Isabella Poggi.
\newblock A {{Model}} of {{Attention}} and {{Interest Using Gaze Behavior}}.
\newblock In \emph{Intelligent {{Virtual Agents}}}, pages 229--240, Berlin,
  Heidelberg, 2005. Springer.

\bibitem[Ren et~al.(2023)Ren, Zhang, and Feng]{renGazeEstimationBased2023}
Guojing Ren, Yang Zhang, and Qingjuan Feng.
\newblock Gaze {{Estimation Based}} on {{Attention Mechanism Combined With
  Temporal Network}}.
\newblock \emph{IEEE Access}, 11:\penalty0 107150--107159, 2023.

\bibitem[Shepherd(2010)]{shepherdFollowingGazeGazeFollowing2010}
Stephen~V. Shepherd.
\newblock Following {{Gaze}}: {{Gaze-Following Behavior}} as a {{Window}} into
  {{Social Cognition}}.
\newblock \emph{Frontiers in Integrative Neuroscience}, 4, 2010.

\bibitem[Tan and Le(2020)]{tanEfficientNetRethinkingModel2020}
Mingxing Tan and Quoc~V. Le.
\newblock {{EfficientNet}}: {{Rethinking Model Scaling}} for {{Convolutional
  Neural Networks}}, 2020.

\bibitem[Tan and Le(2021)]{tanEfficientNetV2SmallerModels2021}
Mingxing Tan and Quoc~V. Le.
\newblock {{EfficientNetV2}}: {{Smaller Models}} and {{Faster Training}}, 2021.

\bibitem[Tan et~al.(2020)Tan, Pang, and
  Le]{tanEfficientDetScalableEfficient2020}
Mingxing Tan, Ruoming Pang, and Quoc~V. Le.
\newblock {{EfficientDet}}: {{Scalable}} and {{Efficient Object Detection}},
  2020.

\bibitem[Tono(2011)]{tonoApplicationEyeTrackingEfl2011}
Yukio Tono.
\newblock Application {{Of Eye-Tracking In Efl Learners}}' {{Dictionary Look-Up
  Process Research}}.
\newblock \emph{International Journal of Lexicography}, 24\penalty0
  (1):\penalty0 124--153, 2011.

\bibitem[Vaswani et~al.(2017)Vaswani, Shazeer, Parmar, Uszkoreit, Jones, Gomez,
  Kaiser, and Polosukhin]{vaswaniAttentionAllYou2017}
Ashish Vaswani, Noam Shazeer, Niki Parmar, Jakob Uszkoreit, Llion Jones,
  Aidan~N. Gomez, Lukasz Kaiser, and Illia Polosukhin.
\newblock Attention {{Is All You Need}}, 2017.

\bibitem[Wang et~al.(2020)Wang, Wu, Zhu, Li, Zuo, and
  Hu]{wangECANetEfficientChannel2020}
Qilong Wang, Banggu Wu, Pengfei Zhu, Peihua Li, Wangmeng Zuo, and Qinghua Hu.
\newblock {{ECA-Net}}: {{Efficient Channel Attention}} for {{Deep Convolutional
  Neural Networks}}, 2020.

\bibitem[Wasfy et~al.(2024)Wasfy, Anber, and
  Atia]{wasfyEyeingInterfaceAdvancing2024}
Ahmed Wasfy, Nourhan Anber, and Ayman Atia.
\newblock Eyeing the {{Interface}}: {{Advancing UI}}/{{UX Analytics Through Eye
  Gaze Technology}}.
\newblock In \emph{2024 {{Intelligent Methods}}, {{Systems}}, and
  {{Applications}} ({{IMSA}})}, pages 538--543, 2024.

\bibitem[Yamamoto and
  {Imai-Matsumura}(2013)]{yamamotoTeachersGazeAwareness2013}
Tsuyoshi Yamamoto and Kyoko {Imai-Matsumura}.
\newblock Teachers' {{Gaze}} and {{Awareness}} of {{Students}}' {{Behavior}}:
  {{Using An Eye Tracker}}.
\newblock \emph{Comprehensive Psychology}, 2:\penalty0 01.IT.2.6, 2013.

\bibitem[Zhang et~al.(2015)Zhang, Sugano, Fritz, and
  Bulling]{zhangAppearancebasedGazeEstimation2015a}
Xucong Zhang, Yusuke Sugano, Mario Fritz, and Andreas Bulling.
\newblock Appearance-based gaze estimation in the wild.
\newblock In \emph{2015 {{IEEE Conference}} on {{Computer Vision}} and
  {{Pattern Recognition}} ({{CVPR}})}, pages 4511--4520, Boston, MA, USA, 2015.
  IEEE.

\bibitem[Zhang et~al.(2017)Zhang, Sugano, Fritz, and
  Bulling]{zhangItsWrittenAll2017}
Xucong Zhang, Yusuke Sugano, Mario Fritz, and Andreas Bulling.
\newblock It's {{Written All Over Your Face}}: {{Full-Face Appearance-Based
  Gaze Estimation}}.
\newblock In \emph{2017 {{IEEE Conference}} on {{Computer Vision}} and
  {{Pattern Recognition Workshops}} ({{CVPRW}})}, pages 2299--2308, 2017.

\bibitem[Zhang et~al.(2019)Zhang, Sugano, Fritz, and
  Bulling]{zhangMPIIGazeRealWorldDataset2019}
Xucong Zhang, Yusuke Sugano, Mario Fritz, and Andreas Bulling.
\newblock {{MPIIGaze}}: {{Real-World Dataset}} and {{Deep Appearance-Based Gaze
  Estimation}}.
\newblock \emph{IEEE Transactions on Pattern Analysis and Machine
  Intelligence}, 41\penalty0 (1):\penalty0 162--175, 2019.

\bibitem[Zhang et~al.(2020)Zhang, Park, Beeler, Bradley, Tang, and
  Hilliges]{zhangETHXGazeLargeScale2020a}
Xucong Zhang, Seonwook Park, Thabo Beeler, Derek Bradley, Siyu Tang, and Otmar
  Hilliges.
\newblock {{ETH-XGaze}}: {{A Large Scale Dataset}} for {{Gaze Estimation Under
  Extreme Head Pose}} and {{Gaze Variation}}.
\newblock In \emph{Computer {{Vision}} -- {{ECCV}} 2020}, pages 365--381.
  Springer International Publishing, Cham, 2020.

\bibitem[Zhou et~al.(2019)Zhou, Lin, Jiang, and Chen]{zhouLearning3DGaze2019}
Xiaolong Zhou, Jianing Lin, Jiaqi Jiang, and Shengyong Chen.
\newblock Learning {{A 3D Gaze Estimator}} with {{Improved Itracker Combined}}
  with {{Bidirectional LSTM}}.
\newblock In \emph{2019 {{IEEE International Conference}} on {{Multimedia}} and
  {{Expo}} ({{ICME}})}, pages 850--855, 2019.

\end{thebibliography}
}
\appendix
\section*{Overview}
This document provides supplementary details for our paper. We include the complete results of our ablation study, a more granular performance analysis on the EVE validation set, and a detailed breakdown of our model's complexity.

\section{Full Ablation Study Results}

Table~\ref{tab:ablation_full} presents the complete results of our ablation study on both the EVE validation and test sets. These results provide the empirical evidence for the design choices discussed in the main paper.

\begin{table}[h!]
    \centering
    \caption{Results of our ablation study of \model{}'s architecture on the EVE \textit{validation} and \textit{test} set. Our full model provides the best performance, with the SAM and our recurrence structure being the most critical components for generalization to the test set.}
    \label{tab:ablation_full}
    \begin{tabular}{lcc}
        \toprule
            \textbf{Model Configuration} & \multicolumn{2}{c}{\textbf{Angular Error ($^\circ$)}} \\
        \cmidrule(lr){2-3} & \textbf{Validation} & \textbf{Test} \\
        \midrule
            \textbf{\model{} (Full Model)} & \textbf{1.86} & \textbf{2.58} \\
        \midrule
            \textit{Ablating Core Modules:} \\
            \quad w/o ECA Module & 2.03 & 2.56 \\
            \quad w/o Self-Attention Module & 2.02 & 4.84 \\
            \quad w/o GRU (Static Model) & 2.24 & 2.88 \\
            \quad Spatial Pooling before GRU & 2.00 & 2.79\\
        \midrule
            \textit{Ablating Backbone \& Fusion:} \\
            \quad ResNet-18 Backbone & 2.25 & 4.22 \\
            \quad EfficientNet-B0 Backbone & 2.10 & 2.78 \\
            \quad EfficientNet-B7 Backbone & 2.12 & 2.90 \\
            \quad EfficientNet-V2-S Backbone & 2.69 & 3.18 \\
            \quad Early Fusion Strategy & 2.26 & 3.26 \\
        \bottomrule
    \end{tabular}
\end{table}

The results in Table~\ref{tab:ablation_full} confirm the trends discussed in the main paper, while also revealing a critical discrepancy between the validation and test sets. 
The validation set is a useful indicator for many design choices; for instance, it correctly shows that our spatio-temporal recurrence ($1.86^\circ$) is superior to the conventional "Spatial Pooling before GRU" approach ($2.00^\circ$). 
However, the validation set fails to predict the critical importance of certain components for robust generalization. 
Notably, removing the Self-Attention Module or using a ResNet-18 backbone results in only a modest performance drop on the validation set (to $2.02^\circ$ and $2.25^\circ$ respectively). 
In contrast, these same configurations catastrophically fail on the test set ($4.84^\circ$ and $4.22^\circ$). 
This discrepancy strongly suggests that the validation set, while useful for initial tuning, does not fully capture the challenging appearance and pose variations present in the test set, which are necessary to evaluate true "in-the-wild" generalization.

\begin{figure}[h!]
    \centering
    \begin{subfigure}[b]{0.98\linewidth}
        \centering
        \includegraphics[width=\textwidth]{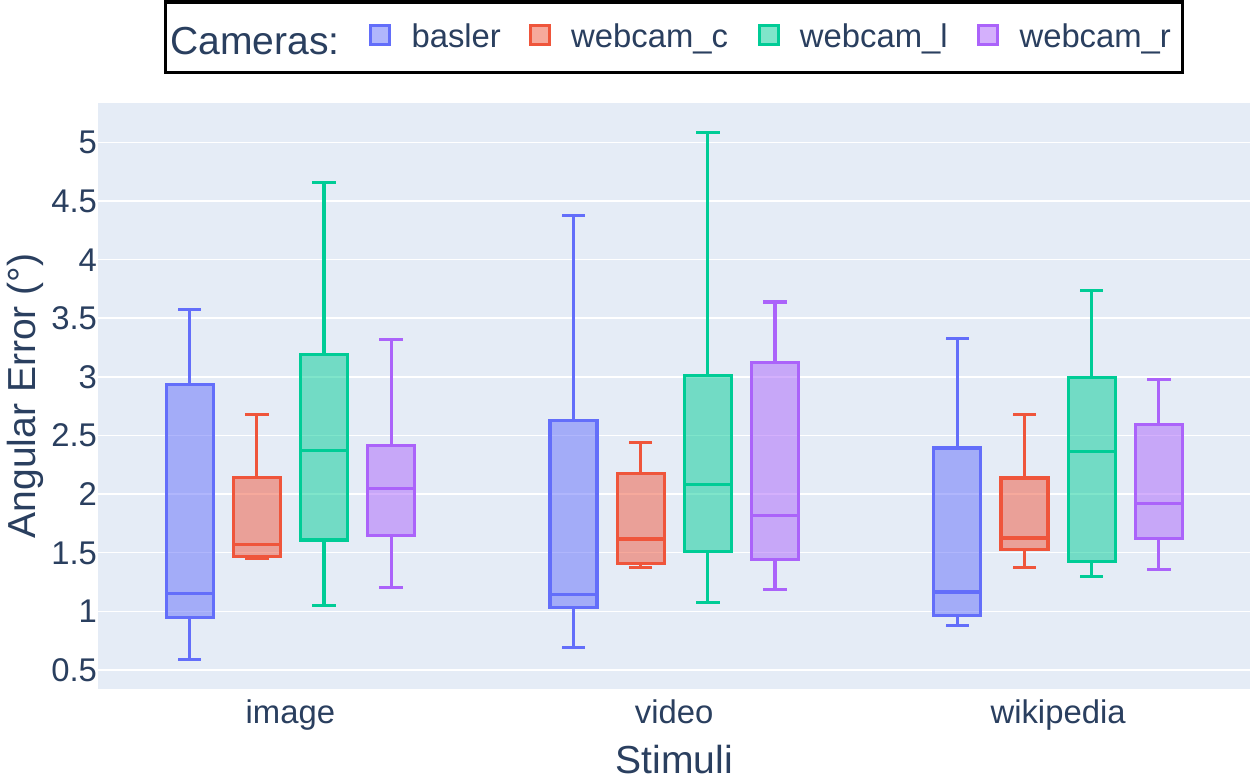}
        \caption{Performance across stimuli and camera views. The central webcam (\textit{webcam\_c}) offers the most stable predictions, while the asymmetry between \textit{webcam\_l} and \textit{webcam\_r} suggests a dataset bias.}
        \label{fig:error_stim_cam}
    \end{subfigure}
    
    \vspace{1em}
    
    \begin{subfigure}[b]{0.98\linewidth}
        \centering
        \includegraphics[width=\linewidth]{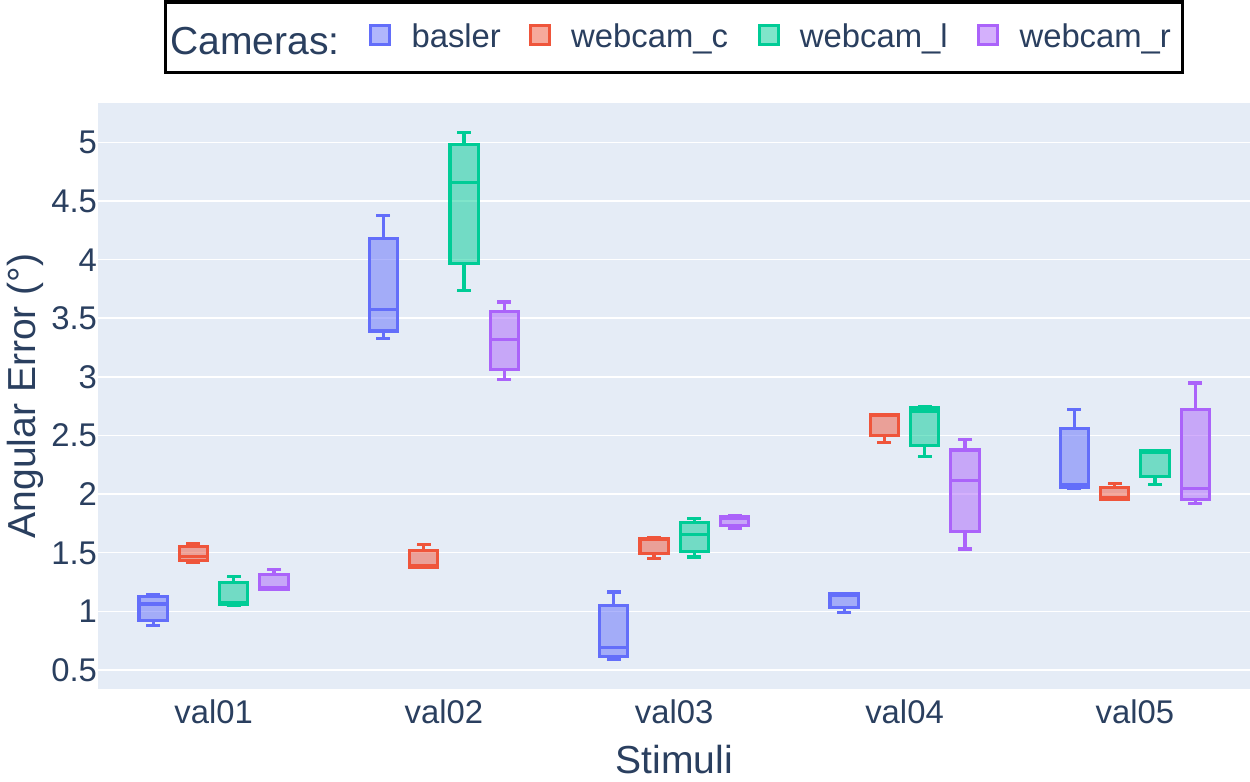}
        \caption{Performance across individual participants. Note the significant variance, with \textit{val02} representing a challenging outlier, highlighting the impact of person-specific factors.}
        \label{fig:error_part_cam}
    \end{subfigure}

    \caption{Detailed performance analysis of \model{} on the EVE validation set. These figures provide a granular breakdown of the aggregate results presented in the main paper's ablation study.}
    \label{fig:combined_errors}
\end{figure}

\section{Detailed Performance Analysis on the EVE Validation Set}

To provide a more granular understanding of our model's behaviour, we present a detailed analysis of its performance on the EVE validation set in Figure~\ref{fig:combined_errors}.

\subsection{Influence of Camera and Stimuli}
Panel~\ref{fig:error_stim_cam} breaks down performance by camera view and stimulus type. 
The model is largely robust to the content being viewed, with only a slight, expected increase in error for dynamic \textit{Video} stimuli. 
A more significant trend emerges across the camera views. While the high-quality, centrally-mounted `basler` camera yields the lowest average error, the \textit{webcam\_c} offers the most consistent predictions with the lowest variance. 
The notable performance gap between the left and right webcams (\textit{webcam\_l} vs. \textit{webcam\_r}) suggests an underlying asymmetry in the dataset, likely from environmental factors such as uneven lighting, which architectural choices like flipping input images cannot fully compensate for.

\subsection{Influence of Participant}
While environmental variations are important, Panel~\ref{fig:error_part_cam} reveals that the most dominant source of performance variance is the individual participant. 
The model performs exceptionally well for participants \texttt{val01} and \texttt{val03}, achieving consistently low average errors, typically between $0.75^\circ$ and $1.75^\circ$, with low deviation across all cameras. 
This demonstrates that our generalized model can achieve very high accuracy for certain individuals.

In stark contrast, participant \texttt{val02} is a notable outlier, with errors ranging from $3.0^\circ$ to $5.0^\circ$ on three of the four cameras. 
Interestingly, performance on the central \textit{webcam\_c} view is substantially better for this participant, suggesting a strong sensitivity to specific head poses or camera angles for certain individuals. 
The remaining participants exhibit more nuanced patterns. For \texttt{val04}, the model is highly accurate with the \textit{basler} camera ($1.0^\circ$ error) but struggles more with the lower-resolution webcams ($2.0^\circ-2.5^\circ$ error), indicating a sensitivity to input image quality. 
Participant \texttt{val05} shows consistent, moderate performance across all views, with an error between $2.0^\circ$ and $2.5^\circ$. 
This detailed per-participant analysis underscores the central challenge of person-independent gaze estimation and motivates the need for the person-specific adaptation methods discussed in the main paper.

\section{Model Complexity}

For completeness, we provide a detailed breakdown of our model's complexity.
Table~\ref{tab:parameter_distribution} details the parameter distribution across the main components of \model{}.

\begin{table}[h!]
    \centering
    \caption{Parameter distribution within our proposed \model{}. The majority of parameters reside in the feature encoders, while our novel attention and recurrence modules are highly parameter-efficient.}    \label{tab:parameter_distribution}
    \begin{tabular}{lcc}
        \toprule
            \textbf{Component} & {\textbf{Parameters}} & {\textbf{\% of Total}} \\
        \midrule
            Eye Encoder & 10 M & 47.37 \\
            Face Encoder & 10 M & 47.19 \\
            ECA Module & 5 & 0.00 \\
            Self-Attention Module & 815 k & 3.80 \\
            Spatio-Temporal GRU & 309 k & 1.44 \\
            Gaze Regression & 41 k & 0.19 \\
        \midrule
            \textbf{Total} & \textbf{21 M} & \textbf{100.00} \\
        \bottomrule
    \end{tabular}
\end{table}

The novel spatio-temporal recurrence module that is central to our contribution comprises less than 2\% of the total parameters, demonstrating its efficiency.
Additionally, the number of parameters in the EfficientNet-B3 encoder ($10$ M) \cite{tanEfficientNetRethinkingModel2020} is similar to ResNet18 ($11$ M) \cite{heDeepResidualLearning2015}, as such, our usage of this encoder does not incur any significant increase in computational cost compared to other dual-input models like FE-NET \cite{renGazeEstimationBased2023} or STTDN \cite{liAppearanceBasedGazeEstimation2023a}.

\end{document}